\pdfoutput=1

\documentclass[11pt]{article}

\usepackage{EMNLP2022}

\usepackage{times}
\usepackage{latexsym}

\usepackage[T1]{fontenc}

\usepackage[utf8]{inputenc}

\usepackage{microtype}

\usepackage{inconsolata}

\usepackage{bbding}
\usepackage{array}
\usepackage{multirow}
\usepackage{graphicx}
\usepackage{comment}
\usepackage{amsfonts}
\usepackage{scalerel}
\usepackage{amsmath,bm}
\usepackage{tabularx}
\usepackage{color}
\usepackage{framed}
\usepackage{booktabs}
\usepackage{arydshln}
\usepackage{tabu}
\usepackage{stfloats}
\usepackage{makecell}
\definecolor{light-gray}{gray}{0.92}

\title{Generative Prompt Tuning for Relation Classification}

\author{
	Jiale Han$^1$, Shuai Zhao$^1$, Bo Cheng$^1$, Shengkun Ma$^1$ \and Wei Lu$^2$
	\\
	\textsuperscript{1}State Key Laboratory of Networking and Switching Technology,\\Beijing University of Posts and Telecommunications\\
	\textsuperscript{2}StatNLP Research Group, Singapore University of Technology and Design\\
	\texttt{\{hanjl,zhaoshuaiby,chengbo,mashengkun\}@bupt.edu.cn, luwei@sutd.edu.sg}
}
\begin{document}
	\maketitle
	\begin{abstract}
		Using prompts to explore the knowledge contained within pre-trained language models for downstream tasks has now become an active topic. Current prompt tuning methods mostly convert the downstream tasks to masked language modeling problems by adding cloze-style phrases and mapping all labels to verbalizations with fixed length, which has proven effective for tasks with simple label spaces.
		However, when applied to relation classification exhibiting complex label spaces, vanilla prompt tuning methods may struggle with label verbalizations with arbitrary lengths due to rigid prompt restrictions.
		Inspired by the text infilling task for pre-training generative models that can flexibly predict missing spans, we propose a novel generative prompt tuning method to reformulate relation classification as an infilling problem, which frees our approach from limitations of current prompt based approaches and thus fully exploits rich semantics of entity and relation types.
		In addition, we design entity-guided decoding and discriminative relation scoring to generate and align relations effectively and efficiently during inference. Extensive experiments under fully supervised settings and low-resource settings demonstrate the effectiveness of our approach.
	\end{abstract}
	
	\section{Introduction}
	
	\begin{figure}[t!]
		\centering
		\includegraphics[width=\linewidth]{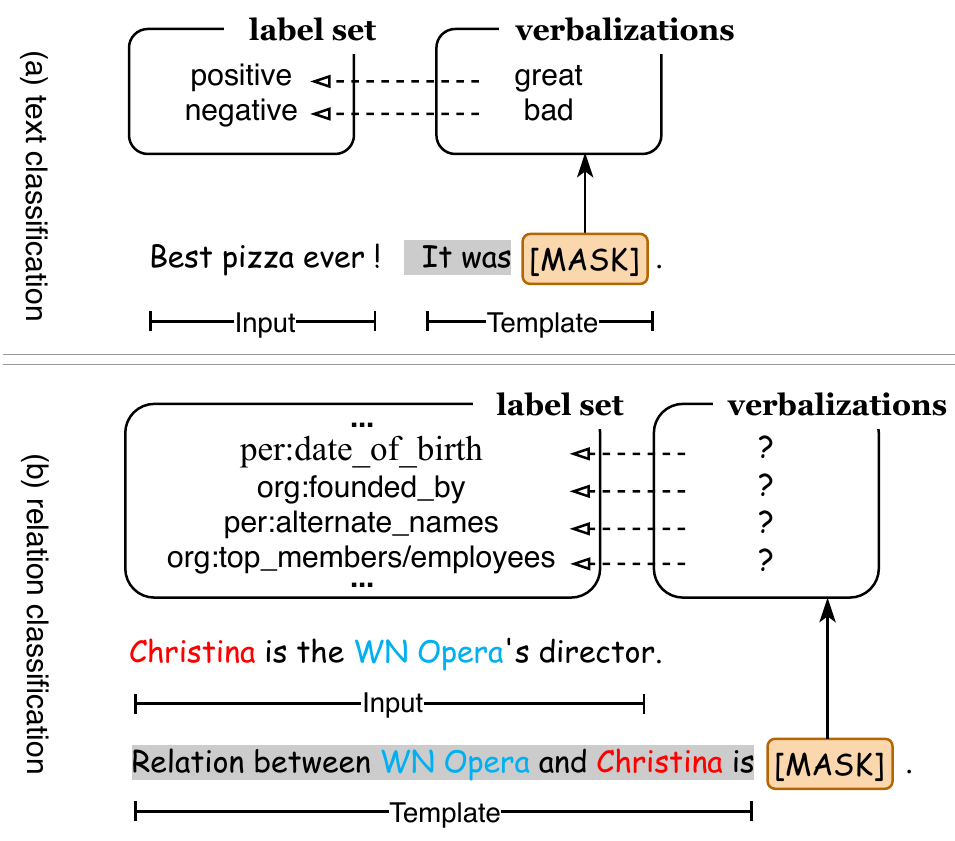}
		\caption{Examples of prompt tuning applied to (a) text classification and (b) relation classification. Existing prompt-based approaches are effective when the label space is simple, but struggle in cases where labels require more complex and elaborate descriptions. We can see from Figure (b) that different classes own label tokens of arbitrary lengths, and it may not always be easy to map them to verbalizations of the same length without losing semantic information.}
		\label{intro} 
	\end{figure}
	Relation classification (RC) is a fundamental task in natural language processing (NLP), aiming to detect the relations between the entities contained in a sentence. With the rise of a series of pre-trained language models (PLMs) \cite{devlin-etal-2019-bert, DBLP:journals/corr/abs-1907-11692, lewis-etal-2020-bart, DBLP:journals/jmlr/RaffelSRLNMZLL20}, fine-tuning PLMs has become a dominating approach to RC \cite{joshi-etal-2020-spanbert, DBLP:conf/aaai/XueSZC21, DBLP:journals/corr/abs-2102-01373}. However, the significant objective gap between pre-training and fine-tuning may hinder the full potential of pre-trained knowledge for such a downstream task.
	
	To this end, prompt tuning \cite{DBLP:conf/nips/BrownMRSKDNSSAA20, schick-schutze-2021-exploiting, schick-schutze-2021-just, DBLP:journals/corr/abs-2107-13586} has been recently proposed and proven effective especially for low-resource scenarios \cite{gao-etal-2021-making, DBLP:conf/naacl/ScaoR21}. The core idea is to convert the objective of downstream tasks to be closer to that of the pre-training tasks, which designs a template to reformulate input examples as cloze-style phrases and a verbalizer to map labels to candidate words. By predicting the mask token, we can determine the label of the input example.

	One disadvantage of prompt tuning is the rigid template restriction, in which the number and position of masked tokens are typically fixed. As presented in Figure~\ref{intro},
	when the label space is simple, downstream tasks can easily adapt to this paradigm \cite{DBLP:conf/acl/HambardzumyanKM20, DBLP:conf/emnlp/LesterAC21}, which predicts one verbalization token at one masked position. However, when applying prompt tuning to RC with complex label space that conveys rich semantic information, vanilla prompt tuning methods may struggle with handling complex label verbalizations with varying lengths. As an attempt to resolve this issue, \citet{DBLP:journals/corr/abs-2105-11259} abridge different labels into verbalizations of fixed length, which, however, may lead to loss of important semantic information. \citet{DBLP:conf/emnlp/SainzLLBA21} convert RC to an entailment problem with hand-crafted verbalizations as hypothesis. Such an approach requires expert efforts, making it difficult to adapt to new datasets and tasks.

	We argue that the fundamental reason for this limitation is that the existing prompt tuning methods imitate masked language modeling (MLM), which predicts only one token at one masked position. Different from MLM, the text infilling task \cite{DBLP:journals/corr/abs-1901-00158} for pre-training generative models \cite{lewis-etal-2020-bart, DBLP:journals/jmlr/RaffelSRLNMZLL20} appears to be more compatible with RC. The task drops consecutive spans of tokens and learns to predict not only which but also how many tokens are missing from each snippet. Following this paradigm allows the model to generate an arbitrary number of tokens at multiple prediction slots.

	This paper proposes a novel \textit{Gen}erative \textit{P}rompt \textit{T}uning method (GenPT), which reformulates RC as a text infilling task to eliminate the rigid prompt restrictions and thus fully exploit the label semantics. 
	Entity type information is further injected thanks to the flexible task format, which is crucial for RC \cite{DBLP:journals/corr/abs-2102-01373}. Specifically, we construct a multi-slot continuous prompt, in which the template converts input sentences to infilling style phrases by leveraging three sentinel tokens as placeholders and desires to fill in the label verbalizations of head entity type, tail entity type, and relation, respectively. Trainable continuous prompt embeddings are employed to avoid manual prompt engineering.
	In addition, how to efficiently determine the final class label is a practical problem when applying generative models to discriminative tasks. 
	We design entity-guided decoding and relation scoring strategies to align the generated sequences with the pre-defined set of labels, making the prediction process more effective and efficient.

	Extensive experiments are conducted on four widely used relation classification datasets under fully supervised and low-resource settings. 
	Compared to a series of strong discriminative and generative baselines, our method achieves better 
	performance, especially in cases where relations are rarely seen during training, demonstrating the effectiveness of our approach. Our main contributions are summarized as follows:\footnote{\textcolor{black}{Our code is available at \url{https://github.com/hanjiale/GenPT}.}}
\begin{itemize}
	\item We reformulate RC as a text infilling task and propose a novel generative prompt tuning method, which eliminates the rigid prompt restrictions and makes full use of semantic information of entity types and relation labels.
	\item We design entity-guided decoding and discriminative relation scoring strategies to predict relations effectively and efficiently.
	\item Experiments on four datasets demonstrate the effectiveness of our model in both fully supervised and low-resource settings.
\end{itemize}

\section{Background}
\begin{figure*}[t!]
	\centering
	\includegraphics[width=0.9\linewidth]{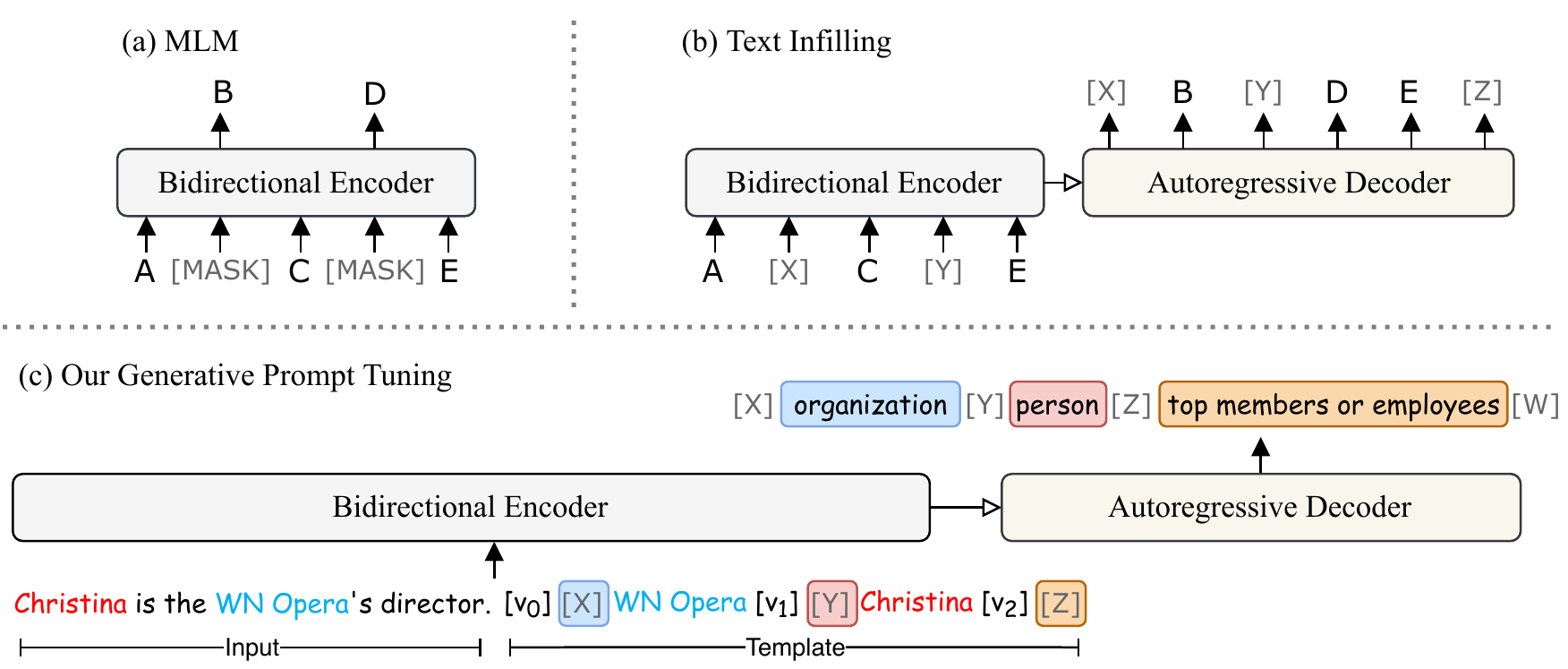}
	\caption{An illustration of (a) MLM pre-training, (b) text infilling pre-training, and (c) our proposed generative prompt tuning approach for RC.}
	\label{model} 
\end{figure*}
\subsection{MLM and Text Infilling}
Masked language modeling \cite{taylor1953cloze} is widely adopted as a pre-training task to obtain a bidirectional pre-trained model \cite{devlin-etal-2019-bert, DBLP:journals/corr/abs-1907-11692, DBLP:conf/nips/ConneauL19}. Generally speaking, a masked language model (MLM) randomly masks out some tokens from the input sentences. Each \texttt{[MASK]} corresponds to one token. The objective is to predict the masked word based on the rest of the tokens (see Figure~\ref{model} (a)). Different from MLM which only predicts one token for one \texttt{[MASK]}, the text infilling task for pre-training seq2seq models \cite{DBLP:journals/jmlr/RaffelSRLNMZLL20, lewis-etal-2020-bart} can flexibly recover spans of different lengths. As shown in Figure~\ref{model} (b), the text infilling task samples a number of text spans with different lengths from the original sentence. Then each span is replaced with a single sentinel token. The encoder is fed with the corrupted sequence, and the decoder sequentially produces the consecutive tokens of dropped-out spans delimited by sentinel tokens. This task is more flexible and can be more compatible with some complex downstream tasks, but is now significantly overlooked.

\subsection{Prompt-Tuning of PLMs}

For standard fine-tuning of classification, the input instance $\boldsymbol{x}$ is converted to a token sequence $\widetilde{\boldsymbol{x}}=\texttt{[CLS]}\, \boldsymbol{x}\, \texttt{[SEP]}$. \textcolor{black}{The model predicts the output classes by feeding the input sequence into PLMs and adding a classifier on top of the \texttt{[CLS]} representations, }which introduces extra parameters and makes it hard to generalize well, especially for low-resource setting. To this end, \textcolor{black}{prompt-tuning is proposed to make the downstream task consistent with the pre-training task.} Current prompt-tuning approaches mainly cast tasks to cloze-style questions to imitate MLM.
Formally, a prompt consists of two key components \cite{schick-schutze-2021-exploiting}, a template and a verbalizer. The template $ T(\cdot)$ reformulates the original input $\boldsymbol{x}$ as a cloze-style phrase $ T(\boldsymbol{x})$  by adding a set of additional tokens and one \texttt{[MASK]} token. The verbalizer $\phi: \mathcal{R}\rightarrow\mathcal{V}$ maps task labels $\mathcal{R}$ to textual tokens $\mathcal{V}$, where $\mathcal{V}$ refers to a set of
label words in the vocabulary of a language model $\mathcal{M}$. In this way, a classification task is transformed into an MLM task:
\begin{equation*}
	\mathrm{P}(r\in\mathcal{R}|\boldsymbol{x})=\mathrm{P}(\texttt{[MASK]}=\phi(r)|T(\boldsymbol{x}))
\end{equation*}

Most prompt tuning methods include one mask token in the template and map each label to one verbalization token to predict classes. Although effective, it is hard to handle tasks with complex label spaces \textcolor{black}{involving labels of varying lengths}.

\section{Approach}\label{4}
As presented in Figure~\ref{model} (c), this paper considers relation classification as a text infilling style task, which takes the sequence $T(\boldsymbol{x})$ processed by the template as source input and outputs a target sequence $\boldsymbol{y}$ to predict relations. The problem definition is formally given in Section~\ref{4.1}. We first introduce how to construct entity-oriented prompts in Section~\ref{4.2}, and then present the model and training objective in Section~\ref{4.3}. The inference details including entity-guided decoding and relation scoring are discussed in Section~\ref{4.4}.

\subsection{Problem Definition}\label{4.1}

Formally, given an instance $\boldsymbol{x}=[x_1, x_2, ..., x_{|\boldsymbol{x}|}]$ with head and tail entity mentions $\boldsymbol{e}_h$ and $\boldsymbol{e}_t$ spanning several tokens in the sequence, as well as entity types $\boldsymbol{t}_h$ and $\boldsymbol{t}_t$, relation classification task is required to predict the relation $r\in\mathcal{R}$ between the entites, where $\mathcal{R}$ is \textcolor{black}{the set of possible relations}. $\boldsymbol{r}$ represents the corresponding label verbalization.
Take a sentence $\boldsymbol{x}=$ ``\textit{Christina is the Washington National Opera's director}'' with relation $r=$ ``\textit{org:top\_members/employees}'' as an example, $\boldsymbol{e}_h$ and $\boldsymbol{e}_t$ are ``\textit{Christina}'' and ``\textit{Washington National Opera}'', and their entity types are ``\textit{organization}'' and ``\textit{person}'' respectively. The relation label verbalization $\boldsymbol{r}=$ ``\textit{top members or employees}'' \footnote{The relation label verbalization $\boldsymbol{r}$ is derived from label $r$, which involves removing attribute words ``\textit{org:}'', discarding symbols of “\_”, and replacing ``/'' with ``\textit{or}''.}. 

\subsection{Entity-Oriented Prompt Construction}\label{4.2}
We design an entity-oriented continuous template $T(\cdot)$ combining entity mentions and type information, which uses a series of learnable continuous tokens \cite{DBLP:journals/corr/abs-2103-10385} as prompts rather than handcrafted token phrases. Specifically, for an input sentence $\boldsymbol{x}$ with two marked entities $\boldsymbol{e}_h$ and $\boldsymbol{e}_t$, instead of utilizing a template with discrete tokens like ``$\boldsymbol{x}.$ \textit{The} \textit{relation} \textit{between} \texttt{[X]} \,$\boldsymbol{e}_h$\;\textit{and}\;  \texttt{[Y]}\; $\boldsymbol{e}_t$ \;\textit{is}\;\texttt{[Z]}.'' which is hand-crafted \textcolor{black}{and it is hard to find the optimal prompt}, we leverage a few learnable continuous tokens to serve as prompts that can be optimized by gradient descent.
\begin{equation*}
\begin{aligned}
	T(\boldsymbol{x})=	\boldsymbol{x}. \; &[v_{0:n_0-1}] \,\texttt{[X]} \,\boldsymbol{e}_h\, [v_{n_0:n_1-1}] \\&\texttt{[Y]}\, \boldsymbol{e}_t \,[v_{n_1:n2-1}] \,\texttt{[Z]}.
\end{aligned}
\end{equation*}
where $[v_i]\in\mathbb{R}^d$ refers to the $i$-{th} continuous token in the template, and \textcolor{black}{$n_0$, $n_1-n_0$, and $n_2-n_1$ are the lengths of these token phrases.} We add three sentinel tokens in the template, where \texttt{[X]} and \texttt{[Y]} in front of entity mentions denote type information of head and tail entities, and \texttt{[Z]} is used to represent relation label tokens. The target sequence then consists of head and tail entity types and label verbalizations, delimited by the sentinel tokens used in the input plus a final sentinel token \texttt{[W]}.
\begin{equation*}
\boldsymbol{y} =\texttt{[X]} \,\boldsymbol{t}_h \, \texttt{[Y]}\,\boldsymbol{t}_t \,\texttt{[Z]} \, \boldsymbol{r} \,\texttt{[W]} 
\end{equation*}

Our prompt is flexible to handle predicted tokens with arbitrary lengths at arbitrary positions, benefiting from the generative text infilling form.

\subsection{Model and Training}\label{4.3}

Given a PLM $\mathcal{M}$ and a template $T(\boldsymbol{x})$ as input, we map $T(\boldsymbol{x})$ into embeddings in which the continuous tokens are mapped to a sequence of continuous vectors,
\begin{equation*}
\begin{aligned}
	&\mathrm{e}(\boldsymbol{x}), h_0, ..., h_{n_0-1}, \mathrm{e}(\texttt{[X]}), \mathrm{e}(\boldsymbol{e}_h), h_{n_0}, ...,\\& h_{n_1-1}, \mathrm{e}(\texttt{[Y]}), \mathrm{e}(\boldsymbol{e}_t), h_{n_1}, ..., h_{n_2-1}, \mathrm{e}(\texttt{[Z]})
\end{aligned}
\end{equation*}
where $\mathrm{e}(\cdot)$ is the embedding layer of $\mathcal{M}$, $h_i\in \mathbb{R}^d$ are trainable embedding tensors with random initialization, $d$ is the embedding dimension of $\mathcal{M}$, and $0 \leq i < n_2$. We feed the input embeddings to the encoder of the model, and obtain hidden representations $\mathbf{h}$ of the sentence:
\begin{equation*}
\mathbf{h} = \textrm{Enc}(T(\boldsymbol{x}))
\end{equation*}

At the $j$-th step of the decoder, the model attends to previously generated tokens $y_{<j}$ and the encoder output $\mathbf{h}$, and then predicts the probability of the next token:
\begin{equation*}
p(y_j|y_{<j}, T(\boldsymbol{x}))=\textrm{Dec}(y_{<j}, \mathbf{h})
\end{equation*}

We train our model by minimizing the negative log-likelihood of label
text $\boldsymbol{y}$ tokens given $T(\boldsymbol{x})$ as input:
\begin{equation*}
\mathcal{L}_{gen}=-\sum_{j=1}^{|\boldsymbol{y}|}\textrm{log}p(y_j|y_{<j}, T(\boldsymbol{x}))
\end{equation*}

\begin{figure}[t!]
\flushleft
\includegraphics[width=0.9\linewidth]{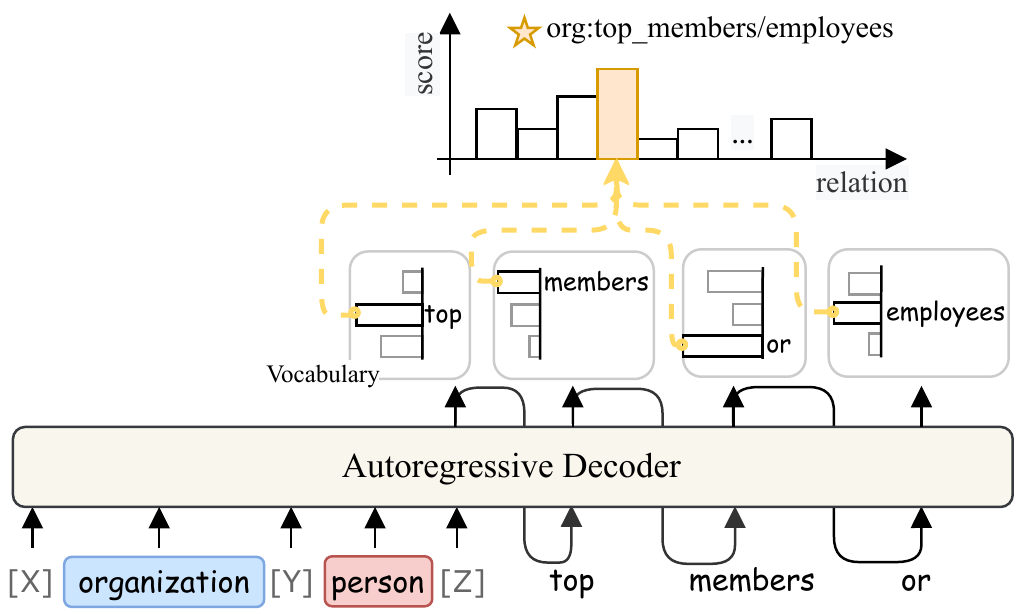}
\caption{Entity-guided decoding and relation scoring.}
\label{decoding} 
\end{figure}
\subsection{Entity-Guided Decoding  and Scoring}\label{4.4}

We propose a simple yet effective entity-guided decoding strategy, which exploits entity type information to implicitly influence the choice of possible candidate relations. As shown in Figure~\ref{decoding}, at the beginning of decoding, instead of only inputting the start-of-sequence token \texttt{<s>} to the decoder, we also append the entity type tokens. 
With $\hat{\boldsymbol{y}}=\texttt{<s>}\,\texttt{[X]}\, \boldsymbol{t}_h \, \texttt{[Y]} \,\boldsymbol{t}_t \,\texttt{[Z]}$ as initial decoder inputs that serves as ``preamble'', the model iteratively predicts the subsequent tokens:
\begin{equation*}
p_{y_j} = p(y_j|\hat{\boldsymbol{y}}, y_{<j}, T(\boldsymbol{x}))
\end{equation*}
{\textcolor{black}{where $p_{y_j}$ is the probability of token $y_j$ at the $j$-th prediction step.
We sum the predicted probabilities of tokens in a label verbalization and normalize it by verbalization length to get the prediction score of the corresponding relation.
Formally, for each relation $r\in \mathcal{R}$ with its label verbalization $\boldsymbol{r}$, the prediction score $s_r$ is calculated as follows:}}
\begin{equation*}
s_r=\frac{1}{|\boldsymbol{r}|}\sum_{j=1}^{\boldsymbol{|r}|}p_{\boldsymbol{r}_j}
\end{equation*}
where $p_{\boldsymbol{r}_j}$ represents the probability of token $\boldsymbol{r}_j$ at the $j$-th step of decoding. 
In this simple way, we can easily align the generated sequences with the label set.
Following the work of \citet{DBLP:conf/emnlp/SainzLLBA21}, we discard relations that do not match the entity types of instances. The sentence is classified into the relation with the highest score.  

\section{Experimental Setup}

\subsection{Datasets and Setups}\label{5.1}

\begin{table*}[t]
\centering
\renewcommand\tabcolsep{3.6pt}
\scalebox{0.7}
{
	\begin{tabular}{llcccccccc}  
		\toprule
		&Model &\textit{Extra Data}&\textit{Entity Type}&{\textit{Label Semantics}}&\textit{Encoder}& TACRED & TACREV & Re-TACRED & Wiki80 \\\midrule
		\multirow{6}*{\rotatebox{90}{{Fine-tuning}}}	&SpanBERT \cite{joshi-etal-2020-spanbert} &\Checkmark&\Checkmark&\XSolidBrush&\small{SpanBERT}&70.8\textcolor{white}{$^\ddag$}&78.0$^\diamondsuit$&85.3$^\P$&88.1$^\star$\\
		&LUKE \cite{yamada-etal-2020-luke}&\Checkmark&\XSolidBrush&\XSolidBrush&\small{LUKE}&72.7\textcolor{white}{$^\ddag$}&80.6$^\ddag$&90.3$^\ddag$&89.2$^\star$\\
		&GDPNet \cite{DBLP:conf/aaai/XueSZC21}&\Checkmark&\Checkmark&\XSolidBrush&\small{SpanBERT}&70.5\textcolor{white}{$^\ddag$}&80.2\textcolor{white}{$^\ddag$}&--\textcolor{white}{$^\ddag$}&--\textcolor{white}{$^\ddag$}\\
		&TANL \cite{DBLP:conf/iclr/PaoliniAKMAASXS21}&\XSolidBrush&\XSolidBrush&\Checkmark&\small{T5}&72.1$^\star$&81.2$^\star$&90.8$^\star$&89.1$^\star$\\
		&NLI \cite{DBLP:conf/emnlp/SainzLLBA21}&\Checkmark&\Checkmark&\Checkmark&\small{RoBERTa-mnli}&	71.0\textcolor{white}{$^\ddag$}&--\textcolor{white}{$^\ddag$}&--\textcolor{white}{$^\ddag$}&--\textcolor{white}{$^\ddag$}\\
		&TYP Marker \cite{DBLP:journals/corr/abs-2102-01373}&\XSolidBrush&\Checkmark&\XSolidBrush&\small{RoBERTa}&74.6\textcolor{white}{$^\ddag$}&83.2\textcolor{white}{$^\ddag$}&91.1\textcolor{white}{$^\ddag$}&89.9$^\star$\\\midrule
		\multirow{5}*{\rotatebox{90}{{Prompt Tuning}}}	
		&PTR \cite{DBLP:journals/corr/abs-2105-11259}&\XSolidBrush&\Checkmark&\Checkmark&\small{RoBERTa}&72.4\textcolor{white}{$^\ddag$}&81.4\textcolor{white}{$^\ddag$}&90.9\textcolor{white}{$^\ddag$}&--\textcolor{white}{$^\ddag$}\\
		&	KnowPrompt \cite{DBLP:journals/corr/abs-2104-07650} &\XSolidBrush&\Checkmark&\Checkmark&\small{RoBERTa}&72.4\textcolor{white}{$^\ddag$}&82.4\textcolor{white}{$^\ddag$}&\textbf{91.3}\textcolor{white}{$^\ddag$}&89.0$^\star$\\
		\cmidrule{2-10}
		&	\textbf{GenPT (BART) }&\XSolidBrush&\Checkmark&\Checkmark&\small{BART}&74.6\textcolor{white}{$^\ddag$}&82.9\textcolor{white}{$^\ddag$}&91.0\textcolor{white}{$^\ddag$}&\textbf{90.8}\textcolor{white}{$^\ddag$}\\
		&	\textbf{GenPT (T5)} &\XSolidBrush&\Checkmark&\Checkmark&\small{T5}&\textbf{75.3}\textcolor{white}{$^\ddag$}&\textbf{84.0}\textcolor{white}{$^\ddag$}&91.0\textcolor{white}{$^\ddag$}&90.6\textcolor{white}{$^\ddag$}\\
		&	\textbf{GenPT (RoBERTa)} &\XSolidBrush&\Checkmark&\Checkmark&\small{RoBERTa}&74.7\textcolor{white}{$^\ddag$}&83.4\textcolor{white}{$^\ddag$}&91.1\textcolor{white}{$^\ddag$}&90.7\textcolor{white}{$^\ddag$}\\
		\bottomrule
	\end{tabular}
}
\caption{Fully supervised results of micro $F_1$ (\%) on four datasets. $\diamondsuit$ are reported by \citet{alt-etal-2020-tacred}, $\P$ are reported by \citet{DBLP:conf/aaai/StoicaPP21}, $\ddag$ are reported by \citet{DBLP:journals/corr/abs-2102-01373}, $\dag$ are reported by \citet{DBLP:journals/corr/abs-2104-07650}, $\star$ indicates we rerun original code, and others are from the original papers. \textbf{Best} numbers are highlighted in each column.}
\label{main_full}
\end{table*}
\begin{figure}[t!]
\centering
\includegraphics[width=0.9\linewidth]{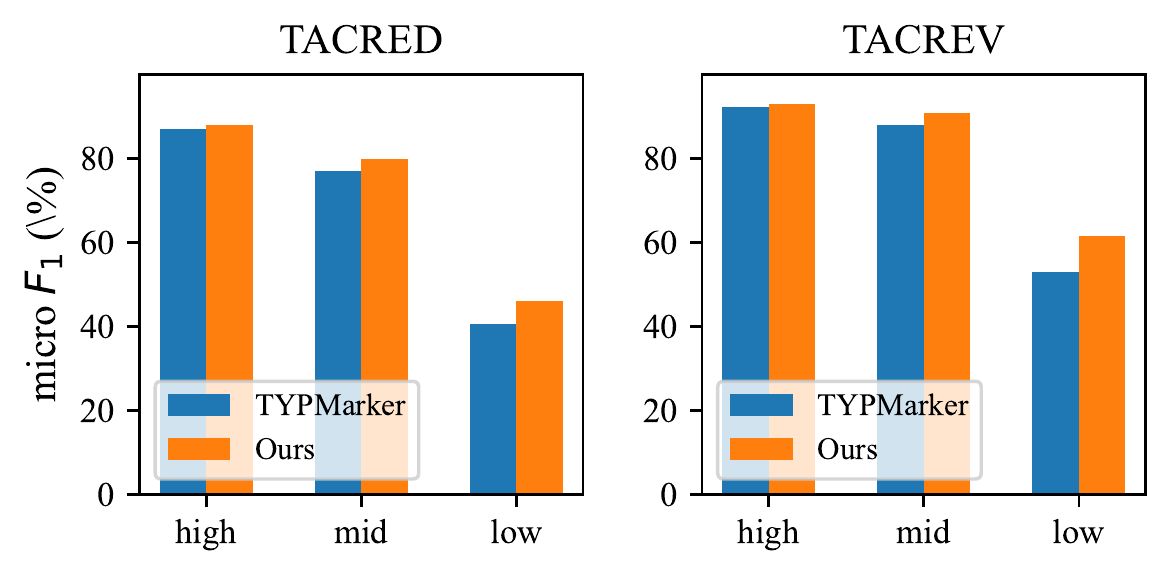}
\caption{Comparison of $F_1$ (\%) with different frequency relations on TACRED and TACREV. See Appendix~\ref{E.2} for detailed scores.}
\label{class_freq} 
\end{figure}

Following the work of \citet{DBLP:journals/corr/abs-2105-11259, DBLP:journals/corr/abs-2104-07650, DBLP:journals/corr/abs-2102-01373}, we conduct experiments on popular RC datasets TACRED \cite{zhang-etal-2017-position}, TACREV \cite{alt-etal-2020-tacred}, and Re-TACRED \cite{DBLP:conf/aaai/StoicaPP21}. Wiki80 \cite{han-etal-2019-opennre} is also adopted to evaluate our proposed model. More details on datasets can be found in Appendix~\ref{A}. We evaluate our model under the fully supervised setting and low-resource setting. For the fully supervised setting, all training and validation data are available. For the low-resource setting, we randomly sample $K$ instances per relation for training and validation, with $K$ being $8$, $16$, and $32$, respectively. Following the work of \citet{gao-etal-2021-making}, we measure the average performance across five different randomly sampled data points using a fixed set of seeds for each experiment. 

\subsection{Implementation Details}
The approach is  based on Pytorch \cite{DBLP:conf/nips/PaszkeGMLBCKLGA19} and the Transformer library of Huggingface \cite{wolf-etal-2020-transformers}. We implement our method on pre-trained generative models T5-large \cite{DBLP:journals/jmlr/RaffelSRLNMZLL20} and BART-large \cite{lewis-etal-2020-bart}. 
For a fair comparison, we further adopt the same pre-trained model RoBERTa-large \cite{DBLP:journals/corr/abs-1907-11692} as in prior work. The implementation details for different pre-trained models are in Appendix~\ref{B}. The training procedures and hyperparameter selection are in Appendix~\ref{C}. We conduct ablation experiments and performance analysis based on the BART version for time efficiency.

\subsection{Baselines}

\textcolor{black}{We compare our model with some recent efforts.}
\noindent\textbf{ \textit{Discriminative fine-tuning methods:}} 1) SpanBERT \cite{joshi-etal-2020-spanbert}, a span-based pre-trained model, 2) LUKE \cite{yamada-etal-2020-luke}, pre-trained contextualized representations of words and entities based on Transformer, 3) GDPNet \cite{DBLP:conf/aaai/XueSZC21}, 
constructing a latent multi-view graph to find indicative words from long sequences for RC, 4) TYP Marker \cite{DBLP:journals/corr/abs-2102-01373}, incorporating entity representations with typed markers, 5) NLI \cite{DBLP:conf/emnlp/SainzLLBA21}, reformulating RC as an entailment problem with hand-crafted verbalizations. 

\noindent\textbf{ \textit{Generative fine-tuning methods:} } 6) TANL \cite{DBLP:conf/iclr/PaoliniAKMAASXS21}, a generative model framing structured prediction as a translation task. We rerun the original code 
based on T5-large.

\noindent\textbf{ \textit{Discriminative prompt-tuning methods:}} 
7) PTR \cite{DBLP:journals/corr/abs-2105-11259}, a prompt tuning method applying logic rules to construct prompts with
several sub-prompts, and 8) KnowPrompt \cite{DBLP:journals/corr/abs-2104-07650}, a knowledge-aware prompt tuning approach that injects knowledge into template design and answer construction. 

Among these works, SpanBERT, LUKE, GDPNet, and NLI adopt extra data for pre-training. SpanBERT, GDPNet, NLI, PTR, KnowPrompt, and TYP Marker utilize entity type information in their methods. TANL, NLI, PTR, and KnowPrompt exploit label semantics. This information is shown in Table~\ref{main_full}.

\section{Results and Discussion}

In this section, we conduct extensive experiments and answer the following questions: 
1) How does our proposed method perform? We demonstrate the superiority of our approach under extensive datasets and settings.
2) What is the impact of prompt design? We conduct extensive experiments to investigate the influence of different prompts.
3) How much benefit can label semantics bring to the RC task? We carry out experiments to illustrate the impact of label semantics on the performance of RE. 
4) What is the effect of the proposed decoding strategy? We perform ablation and case studies to prove the effectiveness and efficiency of decoding strategies.

\subsection{Comparison to Baselines}

\paragraph{Results of fully supervised RC}

\begin{table*}[thbp]
\centering
\scalebox{0.7}{
	\begin{tabular}{llcccccccccccc}  
		\toprule
		&	\multirow{2}*{Model} & 	
		\multicolumn{3}{c}{TACRED} & \multicolumn{3}{c}{TACREV} & \multicolumn{3}{c}{Re-TACRED} & \multicolumn{3}{c}{Wiki80}\\ 
		&&$K$=8& $K$=16& $K$=32 &$K$=8& $K$=16& $K$=32 &$K$=8& $K$=16& $K$=32&$K$=8& $K$=16& $K$=32  \\\midrule
		\multirow{6}*{\rotatebox{90}{{Fine-tuning}}}	&SpanBERT$^\star$&\textcolor{white}{0}8.4\textcolor{white}{$^\ddag$}&17.5\textcolor{white}{$^\ddag$}&17.9\textcolor{white}{$^\ddag$}&\textcolor{white}{0}5.2\textcolor{white}{$^\ddag$}&\textcolor{white}{0}5.7\textcolor{white}{$^\ddag$}&18.6\textcolor{white}{$^\ddag$}&14.2\textcolor{white}{$^\ddag$}&29.3\textcolor{white}{$^\ddag$}&43.9\textcolor{white}{$^\ddag$}&40.2\textcolor{white}{$^\ddag$}&70.2\textcolor{white}{$^\ddag$}&73.6\textcolor{white}{$^\ddag$}\\
		&	LUKE$^\star$ &\textcolor{white}{0}9.5\textcolor{white}{$^\ddag$}&21.5\textcolor{white}{$^\ddag$}&28.7\textcolor{white}{$^\ddag$}&\textcolor{white}{0}9.8\textcolor{white}{$^\ddag$}&22.0\textcolor{white}{$^\ddag$}&29.3\textcolor{white}{$^\ddag$}&14.1\textcolor{white}{$^\ddag$}&37.5\textcolor{white}{$^\ddag$}&52.3\textcolor{white}{$^\ddag$}&53.9\textcolor{white}{$^\ddag$}&71.6\textcolor{white}{$^\ddag$}&81.2\textcolor{white}{$^\ddag$}\\
		&	GDPNet$^\dag$ & 11.8\textcolor{white}{$^\ddag$}&22.5\textcolor{white}{$^\ddag$}&28.8\textcolor{white}{$^\ddag$}&\textcolor{white}{0}8.3\textcolor{white}{$^\ddag$}&20.8\textcolor{white}{$^\ddag$}&28.1\textcolor{white}{$^\ddag$}&18.8\textcolor{white}{$^\ddag$}&48.0\textcolor{white}{$^\ddag$}&54.8\textcolor{white}{$^\ddag$}&45.7\textcolor{white}{$^\ddag$}&61.2\textcolor{white}{$^\ddag$}&72.3\textcolor{white}{$^\ddag$}\\
		&	TANL$^\star$ & 18.1\textcolor{white}{$^\ddag$}&27.6\textcolor{white}{$^\ddag$}&32.1\textcolor{white}{$^\ddag$}&18.6\textcolor{white}{$^\ddag$}&28.8\textcolor{white}{$^\ddag$}&32.2\textcolor{white}{$^\ddag$}&26.7\textcolor{white}{$^\ddag$}&50.4\textcolor{white}{$^\ddag$}&59.2\textcolor{white}{$^\ddag$}&68.5\textcolor{white}{$^\ddag$}&77.9\textcolor{white}{$^\ddag$}&82.2\textcolor{white}{$^\ddag$}\\
		&NLI$^\star$& 31.8\textcolor{white}{$^\ddag$}&35.6\textcolor{white}{$^\ddag$}&36.7\textcolor{white}{$^\ddag$}&--\textcolor{white}{$^\ddag$}&--\textcolor{white}{$^\ddag$}&--\textcolor{white}{$^\ddag$}&--\textcolor{white}{$^\ddag$}&--\textcolor{white}{$^\ddag$}&--\textcolor{white}{$^\ddag$}&--\textcolor{white}{$^\ddag$}&--\textcolor{white}{$^\ddag$}&--\textcolor{white}{$^\ddag$}\\
		&	TYP Marker$^\ddag$ &28.9\textcolor{white}{$^\ddag$} &32.0\textcolor{white}{$^\ddag$}&32.4\textcolor{white}{$^\ddag$}&27.6\textcolor{white}{$^\ddag$}&31.2\textcolor{white}{$^\ddag$}&32.0\textcolor{white}{$^\ddag$}&44.8\textcolor{white}{$^\ddag$}&54.1\textcolor{white}{$^\ddag$}&60.0\textcolor{white}{$^\ddag$}&70.6$^\star$&77.6$^\star$&82.2$^\star$\\\midrule
		\multirow{9}*{\rotatebox{90}{{Prompt Tuning}}}	
		&	PTR $^\ddag$&28.1\textcolor{white}{$^\ddag$}&30.7\textcolor{white}{$^\ddag$}&32.1\textcolor{white}{$^\ddag$}&28.7\textcolor{white}{$^\ddag$}&31.4\textcolor{white}{$^\ddag$}&32.4\textcolor{white}{$^\ddag$}&51.5\textcolor{white}{$^\ddag$}&56.2\textcolor{white}{$^\ddag$}&62.1\textcolor{white}{$^\ddag$}&--\textcolor{white}{$^\ddag$}&--\textcolor{white}{$^\ddag$}&--\textcolor{white}{$^\ddag$}\\
		&	KnowPrompt $^\dag$  &32.0\textcolor{white}{$^\ddag$}&35.4\textcolor{white}{$^\ddag$}&36.5\textcolor{white}{$^\ddag$}&32.1\textcolor{white}{$^\ddag$}&33.1\textcolor{white}{$^\ddag$}&34.7\textcolor{white}{$^\ddag$}&55.3\textcolor{white}{$^\ddag$}&\textbf{63.3}\textcolor{white}{$^\ddag$}&65.0\textcolor{white}{$^\ddag$}&78.5$^\star$&82.2$^\star$&84.5$^\star$\\
		\cmidrule{2-14}
		&	\multirow{2}*{\textbf{GenPT (BART)}}   &32.6\textcolor{white}{$^\ddag$}&35.6\textcolor{white}{$^\ddag$}&37.1\textcolor{white}{$^\ddag$}&31.1\textcolor{white}{$^\ddag$}&34.3\textcolor{white}{$^\ddag$}&\textbf{36.6}\textcolor{white}{$^\ddag$}&52.9\textcolor{white}{$^\ddag$}&58.1\textcolor{white}{$^\ddag$}&63.9\textcolor{white}{$^\ddag$}&77.8\textcolor{white}{$^\ddag$}&82.4\textcolor{white}{$^\ddag$}&85.1\textcolor{white}{$^\ddag$}\\
		&	&\small{$(\pm0.5)$\textcolor{white}{$^\ddag$}}&\small{$(\pm1.8)$\textcolor{white}{$^\ddag$}}&\small{$(\pm1.0)$\textcolor{white}{$^\ddag$}}&\small{$(\pm0.8)$\textcolor{white}{$^\ddag$}}&\small{$(\pm1.9)$\textcolor{white}{$^\ddag$}}&\small{$(\pm1.1)$\textcolor{white}{$^\ddag$}}&\small{$(\pm1.5)$\textcolor{white}{$^\ddag$}}&\small{$(\pm1.8)$\textcolor{white}{$^\ddag$}}&\small{$(\pm1.5)$\textcolor{white}{$^\ddag$}}&\small{$(\pm0.5)$\textcolor{white}{$^\ddag$}}&\small{$(\pm0.7)$\textcolor{white}{$^\ddag$}}&\small{$(\pm0.4)$\textcolor{white}{$^\ddag$}}\\
		&	\multirow{2}*{\textbf{GenPT (T5)}}   &31.8\textcolor{white}{$^\ddag$}&33.3\textcolor{white}{$^\ddag$}&36.0\textcolor{white}{$^\ddag$}&31.3\textcolor{white}{$^\ddag$}&32.6\textcolor{white}{$^\ddag$}&34.6\textcolor{white}{$^\ddag$}&54.7\textcolor{white}{$^\ddag$}&58.7\textcolor{white}{$^\ddag$}&62.5\textcolor{white}{$^\ddag$}&77.1\textcolor{white}{$^\ddag$}&82.0\textcolor{white}{$^\ddag$}&84.3\textcolor{white}{$^\ddag$}\\
		&		&\small{$(\pm0.8)$\textcolor{white}{$^\ddag$}}&\small{$(\pm1.3)$\textcolor{white}{$^\ddag$}}&\small{$(\pm1.1)$\textcolor{white}{$^\ddag$}}&\small{$(\pm1.7)$\textcolor{white}{$^\ddag$}}&\small{$(\pm1.5)$\textcolor{white}{$^\ddag$}}&\small{$(\pm1.2)$\textcolor{white}{$^\ddag$}}&\small{$(\pm2.3)$\textcolor{white}{$^\ddag$}}&\small{$(\pm2.4)$\textcolor{white}{$^\ddag$}}&\small{$(\pm1.0)$\textcolor{white}{$^\ddag$}}&\small{$(\pm0.9)$\textcolor{white}{$^\ddag$}}&\small{$(\pm0.6)$\textcolor{white}{$^\ddag$}}&\small{$(\pm0.4)$\textcolor{white}{$^\ddag$}}\\
		&	\multirow{2}*{\textbf{GenPT (RoBERTa)}}   &\textbf{35.7}\textcolor{white}{$^\ddag$}&\textbf{36.6}\textcolor{white}{$^\ddag$}&\textbf{37.4}\textcolor{white}{$^\ddag$}&\textbf{34.4}\textcolor{white}{$^\ddag$}&\textbf{34.6}\textcolor{white}{$^\ddag$}&36.2\textcolor{white}{$^\ddag$}&\textbf{57.1}\textcolor{white}{$^\ddag$}&60.4\textcolor{white}{$^\ddag$}&\textbf{65.2}\textcolor{white}{$^\ddag$}&\textbf{78.6}\textcolor{white}{$^\ddag$}&\textbf{83.3}\textcolor{white}{$^\ddag$}&\textbf{85.7}\textcolor{white}{$^\ddag$}\\
		&		&\small{$(\pm1.1)$\textcolor{white}{$^\ddag$}}&\small{$(\pm1.5)$\textcolor{white}{$^\ddag$}}&\small{$(\pm1.8)$\textcolor{white}{$^\ddag$}}&\small{$(\pm0.8)$\textcolor{white}{$^\ddag$}}&\small{$(\pm1.6)$\textcolor{white}{$^\ddag$}}&\small{$(\pm1.8)$\textcolor{white}{$^\ddag$}}&\small{$(\pm2.3)$\textcolor{white}{$^\ddag$}}&\small{$(\pm1.3)$}&\small{$(\pm2.0)$\textcolor{white}{$^\ddag$}}&\small{$(\pm0.4)$\textcolor{white}{$^\ddag$}}&\small{$(\pm0.3)$\textcolor{white}{$^\ddag$}}&\small{$(\pm0.4)$\textcolor{white}{$^\ddag$}}\\
		\bottomrule
	\end{tabular}
}
\caption{Low-resource results on four datasets. We report the mean and standard deviation performance of micro $F_1$ (\%) over 5 different splits. 
	Results marked with $\dag$ are reported by \citet{DBLP:journals/corr/abs-2104-07650}, $\ddag$  are reported by \citet{DBLP:journals/corr/abs-2105-11259}, and $\star$ indicates we rerun original code under low-resource settings. \textbf{Best} numbers are highlighted in each column.}
\label{main_low}
\end{table*}

As shown in Table~\ref{main_full}, we evaluate our model in the fully supervised setting. Specifically, our model outperforms the state-of-the-art discriminative fine-tuning model TYP Marker and prompt tuning methods PTR and KnowPrompt. Compared to the generative approach TANL which converts RC to augmented natural language translation, GenPT-T5 improves $3.2$ points in terms of $F_1$ on TACRED, \textcolor{black}{suggesting the format of text infilling can better exploit pre-trained knowledge in language models}.
\paragraph{Impact of training relation frequency}

To further explore the impact of relation frequencies in training data, we split the test set into three subsets according to the class frequency in training. Specifically, we regard the relations with more than $300$ training instances to form a  high frequency subset, those with $50$-$300$ training instances form a middle frequency subset, and the rest form a low frequency subset. 
Detailed statistics and data splits are in Appendix~\ref{E.2}. As shown in Figure~\ref{class_freq}, we evaluate our model and TYP Marker on the three subsets of the test data. Our model outperforms TYP Marker on all three subsets, especially on the low frequency set, proving its effectiveness when the class rarely appears in the training data.

\paragraph{Results of low-resource RC}
Table~\ref{main_low} presents the results of micro $F_1$ under the low-resource setting. Our model achieves the best performance in comparison to existing approaches, proving that our method can better handle extremely few-shot classification tasks. The performance gain mainly comes from reformulating RC to a text infilling task to eliminate the rigid prompt restrictions and thus fully leveraging the label semantics. In addition, we notice our method achieves strong performance based on BART and T5, while yielding the best results in low-resource settings when using the same backbone RoBERTa as the discriminative baselines, which on the one hand fairly demonstrates the effectiveness of our method, on the other hand suggests RoBERTa is more data-efficient.

\subsection{The Effect of Prompt}\label{5.4}
\begin{table*}[t!]
\centering
\renewcommand\tabcolsep{2.pt}
\scalebox{0.7}
{	\begin{tabular}{lllccc}
		\toprule
		\multirow{2}*{No.} & \multirow{2}*{Inputs} &\multirow{2}*{Targets} &	\multicolumn{3}{c}{TACRED} \\ 
		&&&$K$=8&$K$=16&$K$=32\\\midrule
		
		1 & $\boldsymbol{x}.\; [v_{0:n_0-1}] \,\texttt{[MASK]} \,\boldsymbol{e}_h\, [v_{n_0:n_1-1}]\,\texttt{[MASK]} \,\boldsymbol{e}_t \,[v_{n_1:n_2-1}] \,\texttt{[MASK]}$.&\colorbox{white}{\texttt{[MASK]}\,$\boldsymbol{t}_h \, \texttt{[MASK]}\,\boldsymbol{t}_t$\,\texttt{[MASK]}}$\boldsymbol{r}$\,\texttt{[MASK]} &	\textbf{32.6}&\textbf{35.6}&\textbf{37.1}\\\midrule
		2 & $\boldsymbol{x}.$ \textit{The} \textit{relation} \textit{between} \texttt{[MASK]} \,$\boldsymbol{e}_h$\,\textit{and}\,  \texttt{[MASK]}\,$\boldsymbol{e}_t$ \,\textit{is}\;\texttt{[MASK]}.&\colorbox{white}{\texttt{[MASK]}\,$\boldsymbol{t}_h \, \texttt{[MASK]}\,\boldsymbol{t}_t$\,\texttt{[MASK]}}$\boldsymbol{r}$\,\texttt{[MASK]} &	32.3&34.7&37.0\\
		3	&$\boldsymbol{x}.\;  [v_{0:n_0-1}] \,\boldsymbol{t}_h\; \boldsymbol{e}_h\; [v_{n_0:n_1-1}] \,\boldsymbol{t}_t \;\boldsymbol{e}_t \,[v_{n_1:n_2-1}] \,\texttt{[MASK]}$. &\colorbox{white}{\texttt{[MASK]}}$\boldsymbol{r}$\,\texttt{[MASK]}&30.0&33.8&35.6\\
		4 & $\boldsymbol{x}.\; [v_{0:n_0-1}] \, \boldsymbol{e}_h \,[v_{n_0:n_1-1}] \,\boldsymbol{e}_t \,[v_{n_1:n_2-1}] \,\texttt{[MASK]}$.  &\colorbox{white}{\texttt{[MASK]}}$\boldsymbol{r}$\,\texttt{[MASK]}&30.4&34.1&35.9\\
		5 & $\boldsymbol{x}.\; [v_{0:n_2-1}] \,\texttt{[MASK]}$.  &\colorbox{white}{\texttt{[MASK]}}$\boldsymbol{r}$\,\texttt{[MASK]}&23.4&25.2&26.4\\
		6 & $\boldsymbol{x}$.&\;\colorbox{white}{$\boldsymbol{r}$}&22.7&25.0&26.1\\
		\bottomrule
\end{tabular}}

\caption{Ablation study on TACRED showing micro $F_1$ (\%) to illustrate the impact of prompt designs. 
}
\label{ablation}
\end{table*}
\paragraph{The impact of prompt format} 
Extensive experiments with various templates are conducted to illustrate the effect of prompt construction, as shown in Table~\ref{ablation}. We choose to put the relation at the end of the template due to its agreement with autoregressive generation, allowing utilizing entity types as guidance during inference.
Row \#2 displays the results of manually designed template with discrete tokens. 
The results indicate the learnable continuous prompt achieves better performance and frees us from prompt engineering, i.e., picking the optimal discrete template.
As depicted in row \#3, we add entity types to the input sequence instead of predicting them in the targets. Row \#4 represents predicting the relation labels without entity types. 
Row \#5 further removes entity mentions in the template. 
The $F_1$ scores of these implementations all degrade, suggesting the importance of entity types and mentions for RE. Moreover, we compare our model to the vanilla seq2seq fine-tuning method (row \#6). Our model outperforms the vanilla fine-tuning by a large margin, revealing the effectiveness of our entity-oriented prompt design and tuning.
\paragraph{Discussion of continuous token length}

\begin{figure}[t]
\centering
\includegraphics[width=0.9\linewidth]{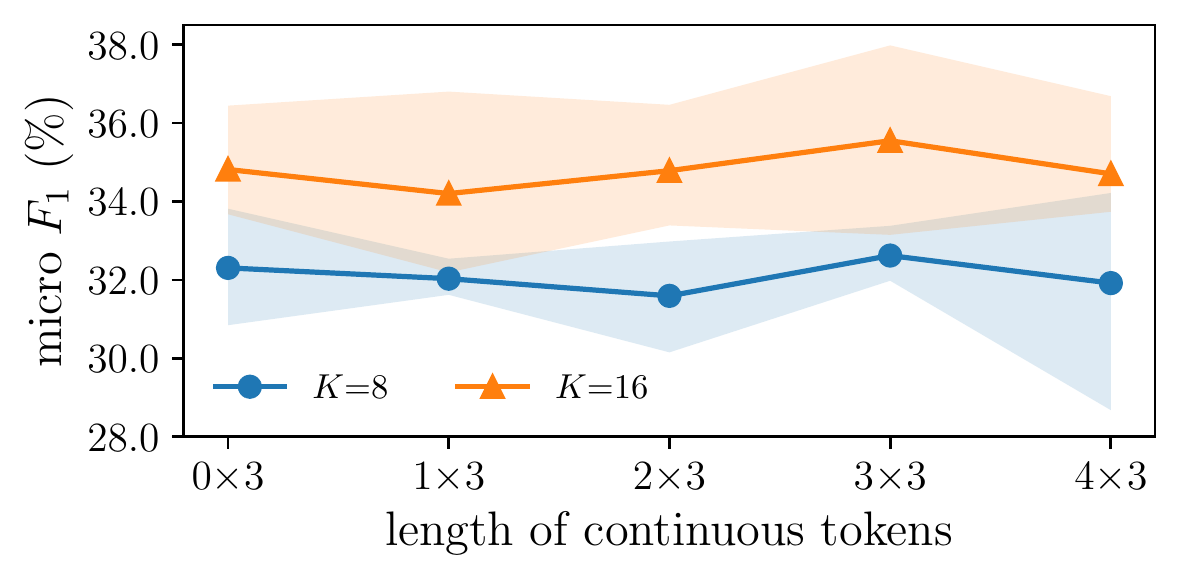}
\caption{Micro $F_1$ (\%) with different numbers of continuous tokens on TACRED.}
\label{prompt_len} 
\end{figure}

Here we discuss the effect of continuous token lengths. The length of continuous tokens in the template $T(\cdot)$ is set to $n\times3$, where $n_0=n_1-n_0=n_2-n_1=n$. The results are shown in Figure~\ref{prompt_len}. The $F_1$ increases when $n$ increases from $0$ to $3$, and then decreases slightly. In our experiment, we fix $n$ to $3$ to achieve effective performance, that is totally $9$ continuous tokens in the template. Interestingly, we observe that the model gains valid performance when $n$ is 0, which suggests our method can be adapted to zero-shot setting by selecting $n$ as $0$. We show the experimental results under zero-shot setting in Appendix~\ref{F}.
\subsection{Analysis of Label Semantics}

Our approach makes full use of label semantics by learning to predict entire label verbalizations with different lengths. To verify the benefits coming from the label semantics, we compare our method with predicting partial-semantic hand-crafted verbalization and non-semantic numeric relation id. The results are presented in Table~\ref{rule}. Specifically, predicting hand-crafted verbalization means mapping each label to manually crafted label verbalization with fixed length, which may lose some semantic information. Here we adopt the hand-crafted verbalizations by the work of \citet{DBLP:journals/corr/abs-2105-11259}.
For example, relation ``\textit{per:country\_of\_birth}'' is mapped to [\textit{person, was, born, in, country}].
See Table~\ref{handmade-labels} in Appendix~\ref{D.3} for all manually crafted label verbalizations. 
For predicting non-semantic numeric relation id, we assign a relation id (e.g., ``\texttt{<Rel\_0>}''), a special text token to a specific relation, and make the model learn to generate the correct relation id. Our model obtains the best results by leveraging full label semantics, proving that label semantics is crucial for the RC task.

\begin{table}[t]
\centering
\renewcommand\tabcolsep{6.pt}
\scalebox{0.7}
{
	\begin{tabular}{lccc}  
		\toprule
		\multirow{2}*{Model} & 	\multicolumn{3}{c}{TACRED} \\ 
		&$K$=8&$K$=16&$K$=32\\\midrule
		Ours (full-semantic)&\textbf{32.6}&\textbf{35.6}&\textbf{37.1}\\
		Handmade verbalization (partial-semantic) &29.8&31.5&33.2\\
		Numeric relation id (non-semantic)&16.2&24.4&30.8\\
		\bottomrule
	\end{tabular}
}
\caption{Analysis of verbalizations with original label tokens, handcrafted tokens with fixed length, and numeric relation ids.}
\label{rule}
\end{table}

\subsection{Analysis of Decoding Strategy}\label{4.6}
\paragraph{The effect of relation scoring} 
When re-running TANL on TACRED, we notice that it takes a long time ($86.62$ hours) to perform inference on the test set. To illustrate the efficiency of our approach, we compare our relation scoring strategy with likelihood-based prediction \cite{nogueira-dos-santos-etal-2020-beyond, DBLP:conf/iclr/PaoliniAKMAASXS21}, which feeds each candidate sequence into decoder and uses output token likelihoods as the corresponding class scores. More details are in Appendix~\ref{D.4}. As shown in Table~\ref{prediction}, our method achieves promising performance with less inference time. In addition, we note all generative methods are slower than their discriminative counterparts due to the autoregressive nature of the process (see Section~\ref{limitations}).

\paragraph{The effect of entity-guided decoding} 
As shown in Table~\ref{case}, by removing the entity-guided decoding, micro $F_1$ drops from $32.6$ to $32.1$ with $K$=$8$, proving its effectiveness. We further carry out a detailed case study. A real test instance from TACRED with its entity type information is also given in Table~\ref{case}. When there is no entity type guidance, the decoder generates the sequence ``\textit{organization location cities of headquarters}'', and incorrectly classifies the instance as relation ``\textit{org:city\_of\_headquarters}''. Our model equipped with entity-guided decoding correctly predicts the relation as ``\textit{org:subsidiaries}''. 
This strategy implicitly restricts the generated candidates, gaining performance improvement.

\begin{table}[t]
\centering
\scalebox{0.7}
{
	\begin{tabular}{lcc}  
		\toprule
		\multirow{2}*{Model} & 	\multicolumn{2}{c}{TACRED} \\ 
		&$K$=8&$K$=16\\\midrule
		Ours&32.6 &35.6 \\
		Likelihood-based Prediction (LP)&32.4&35.5\\
		\midrule
		\multicolumn{3}{c}{$\includegraphics{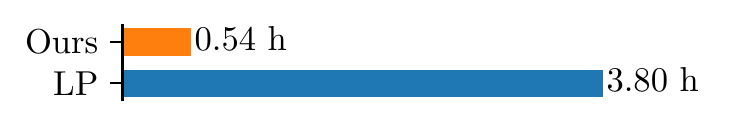}$}\\
		\bottomrule
	\end{tabular}
}
\caption{Micro $F_1$ (\%) and inference time (hours) on the test set with relation scoring and likelihood-based prediction (LP), respectively.}
\label{prediction}
\end{table}

\section{Related Work}

\subsection{Language Model Prompting}
Language model prompting has emerged with the introduction of the GPT series \cite{radford2018improving, radford2019language, DBLP:conf/nips/BrownMRSKDNSSAA20}. PET \cite{schick-schutze-2021-exploiting, schick-schutze-2021-just} reformulates input examples as cloze-style phrases to help language models understand given tasks. 
ADAPET \cite{DBLP:conf/emnlp/TamMBSR21} modifies PET’s objective to provide denser supervision.
However, these methods require manually designed patterns and label verbalizers. To avoid labor-intensive prompt design, automatic prompt search \cite{shin-etal-2020-autoprompt, schick-etal-2020-automatically} has been extensively explored. LM-BFF \cite{gao-etal-2021-making} adopts T5 to generate prompt candidates and verify their effectiveness through prompt tuning.
We would like to highlight the differences between LM-BFF and ours: they adopt T5 for generating templates but still use a single token for each label, whereas we employ T5 to capture variable-length label descriptions. Continuous prompt learning \cite{li-liang-2021-prefix, qin-eisner-2021-learning, DBLP:journals/corr/abs-2103-10385, DBLP:journals/corr/abs-2110-07602, DBLP:conf/iclr/ZhangLCDBTHC22} is further proposed, which directly uses learnable continuous embeddings as prompts rather than discrete token phrases. A few related methods on tasks such as knowledge probing \cite{DBLP:conf/naacl/ZhangFL022} and NER \cite{DBLP:journals/corr/abs-2109-00720} use generative PLMs to generate prompts, or directly generate labels into a sequence without any template, but none of them employs the generative PLMs to fill in the blanks contained in the template as we do.

\begin{table}[t]
\centering
\scalebox{0.7}
{
		\begin{tabular}{lp{1cm}<{\centering}p{1cm}<{\centering}p{1cm}<{\centering}p{1cm}<{\centering}}  
			\toprule
			\multirow{2}*{Model} & 	\multicolumn{4}{c}{TACRED} \\ 
			&$K$=8&$K$=16&$K$=32&full\\\midrule
			Ours&32.6 &35.6 &37.1&74.6\\
			w/o entity-guided decoding&	32.1&35.0&36.7&73.5\\
			\midrule
			\multicolumn{5}{l}{$\includegraphics{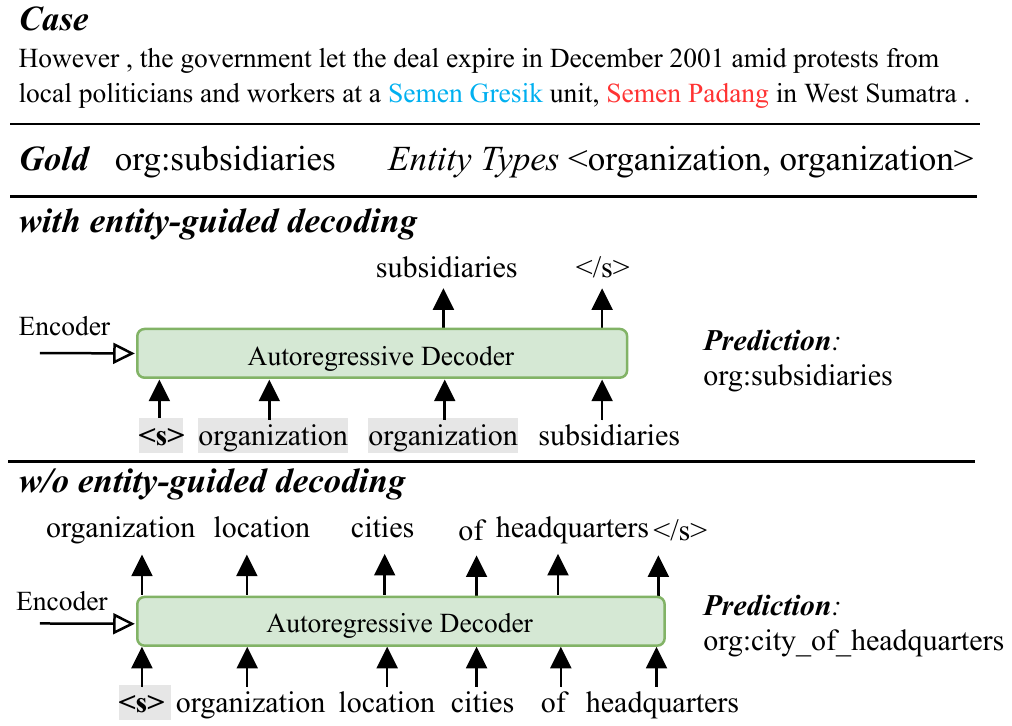}$}\\
			\bottomrule
		\end{tabular}
	}
	\caption{Ablation and case studies to illustrate the effect of entity-guided decoding.}
	\label{case}
\end{table}

\subsection{Relation Classification}
Fine-tuning PLMs for RC  \cite{joshi-etal-2020-spanbert, yamada-etal-2020-luke, DBLP:conf/aaai/XueSZC21, lyu-chen-2021-relation} has achieved promising performance. \citet{DBLP:journals/corr/abs-2102-01373} achieves state-of-the-art results by incorporating entity type information into entity markers. Another interesting line is converting information extraction into generation form. 
\citet{zeng-etal-2018-extracting} and \citet{DBLP:conf/aaai/NayakN20} 
propose seq2seq models to extract relational facts. 
\citet{DBLP:conf/emnlp/HuangTP21} present a generative framework for document-level entity-based 
extraction tasks. There are other works converting RC to translation framework \cite{DBLP:conf/iclr/PaoliniAKMAASXS21, DBLP:conf/emnlp/WangLCH0S21} and entailment task \cite{DBLP:conf/emnlp/SainzLLBA21}. When applying prompt tuning to RC, labels of variable lengths in RC is a major obstacle. To address this issue, existing works \cite{DBLP:journals/corr/abs-2105-11259, DBLP:journals/corr/abs-2104-07650} manually abridge labels into fixed-length tokens, or use one virtual token to represent one relation. Unlike them, GenPT frees prompt tuning from fixed-length verbalizations and predicts complete label tokens. In addition, our method does not need any manual effort, which is more practical when adapted to other datasets or similar tasks.

Few-shot relation classification has attracted recent attention, which can be roughly divided into two directions according to the few-shot setting. One direction is meta-learning based few-shot relation classification \cite{DBLP:conf/emnlp/HanZYWYLS18, DBLP:conf/emnlp/Han0L21, DBLP:conf/cikm/Han0N21}. Given large-scale labeled base class training data and novel class data with scarce labeled instances, the task is required to learn generalized representations from base classes and then adapt these representations to novel classes. The other is generalized few-shot relation classification \cite{DBLP:conf/emnlp/SainzLLBA21, DBLP:journals/corr/abs-2105-11259, DBLP:journals/corr/abs-2104-07650}, which only relies on a small amount of labeled data to train the model. In our work, we conduct experiments following the generalized few-shot setting, which is more practical and challenging.

\section{Conclusion}
This paper presents a novel generative prompt tuning method for RC. Unlike vanilla prompt tuning that converts a specific task into an MLM problem, we reformulate RC as a text infilling task, which can predict label verbalizations with varying lengths at multiple predicted positions and thus better exploit the semantics of entity and relation types. In addition, we design a simple yet effective entity-guided decoding and discriminative scoring strategy to efficiently decode predicted relations, making our generative model more practical. Qualitative and quantitative experiments on four benchmarks prove the effectiveness of our approach. 

Future work includes applying the proposed method to other related tasks where the label space is large with rich semantic information. In addition, we believe GenPT can be adapted to support multiple labels for the same class. This is analogous to the typical NMT model involving many-to-many mapping (where the target may not be unique). How to exactly introduce multiple valid, semantically equivalent label descriptions (including design of label verbalizations, learning process, etc.) is a research topic deserving a separate study.

\section*{Limitations}\label{limitations}
We argue that the main limitation of our work is the time efficiency of the generative approach compared to discriminative methods, even though we are more efficient than other generative models (see Section~\ref{4.6} for a detailed discussion). The decreased time efficiency is mainly due to the autoregressive nature of the generation process. Specifically, compared to the discriminative model PTR based on RoBERTa-large which takes $0.11$ hours to evaluate on TACRED test data (batch size is $32$), our GenPT-BART takes $0.54$ hours with the same inference batch size. GenPT-T5 and GenPT-RoBERTa take more inference time due to larger model parameters and 2-dimensional attention masks respectively, which are $1.97$ hours and $1.29$ hours.

\section*{Acknowledgements}
We would like to thank the anonymous reviewers, our meta-reviewer and senior area chairs for their thoughtful comments and support on this work. We would also like to Shengqiang Zhang for helpful discussions. 
The first author was a visiting PhD student at SUTD when conducting this work. This work is supported by Beijing Nova Program of Science and Technology (Grant No. Z191100001119031), National Natural Science Foundation of China (Grant No.U21A20468, 52071312, 61972043, 61921003), Guangxi Key Laboratory of Cryptography and Information Security (Grant No. GCIS202111), the Open Program of Zhejiang Lab (Grant No.2021PD0AB02), the Fundamental Research Funds for the Central Universities under grant 2020XD-A07-1, and China Scholarship Council Foundation. This research/project is supported in part by the National Research Foundation Singapore and DSO National Laboratories under the AI Singapore Program (AISG Award No: AISG2-RP-2020-016).

\bibliography{custom}
\bibliographystyle{acl_natbib}

\appendix

\section{Datasets}\label{A}

We conduct experiments on four RC datasets, which are TACRED\footnote{\url{https://nlp.stanford.edu/projects/tacred/}} \cite{zhang-etal-2017-position}, TACREV\footnote{\url{https://github.com/DFKI-NLP/tacrev}} \cite{alt-etal-2020-tacred}, Re-TACRED\footnote{\url{https://github.com/gstoica27/Re-TACRED}} \cite{DBLP:conf/aaai/StoicaPP21}, and Wiki80\footnote{\url{https://github.com/thunlp/OpenNRE}} \cite{han-etal-2019-opennre}. Table~\ref{data} summarizes key statistics of the four datasets. 
\paragraph{TACRED} is one of the most widely used RC datasets. Each instance includes a natural sentence sequence, the types and spans of the entity mentions, and the relation held between the entities or  \textit{no\_relation} label if no relation was found.
\paragraph{TACREV} is a dataset revised from TACRED, which has the same training data as the original TACRED and extensively relabeled development and test sets. \paragraph{Re-TACRED} is another completely re-annotated version of TACRED dataset through an improved crowdsourcing strategy. They re-define relation labels to make them more clear and intuitive and re-annotate the full TACRED dataset.
\paragraph{Wiki80} is a relation classification dataset derived from FewRel \cite{han-etal-2018-fewrel}, a large scale few-shot RC dataset. The original data only has a training/development split, so we randomly resplit the training set into a training set and a development set, and treat the original development set as test set, following the split used in \citet{DBLP:journals/corr/abs-2104-07650}. The entity type information is obtained from Wikidata \cite{DBLP:journals/cacm/VrandecicK14}. 

\section{Implementation Details}\label{B}
\begin{figure*}[t]
	\centering
	\includegraphics[width=0.9\textwidth]{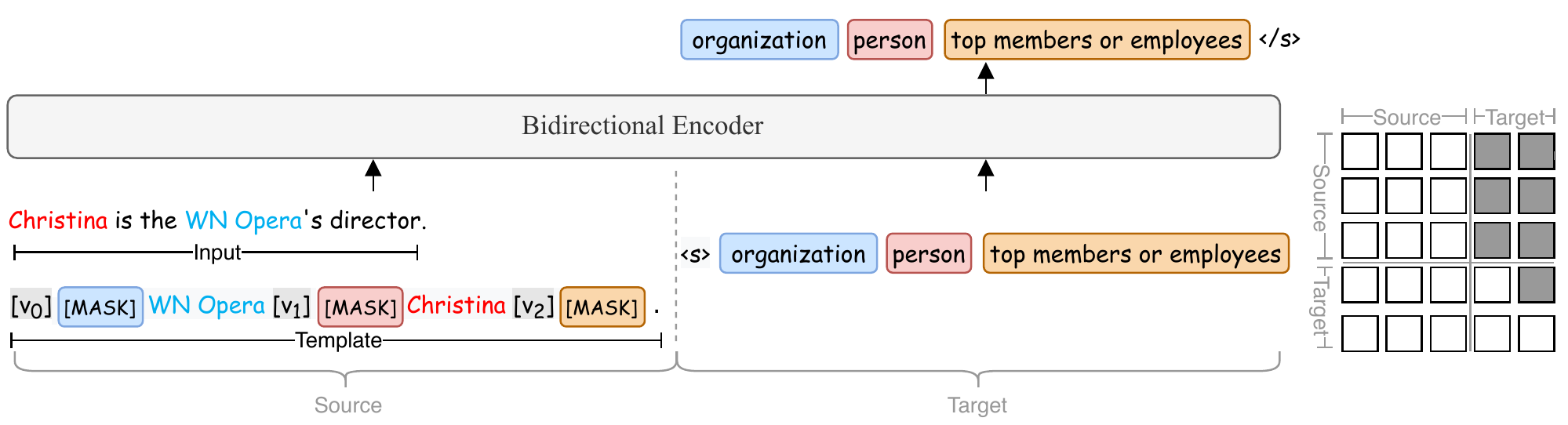}
	\caption{Our proposed generative prompt tuning approach based on RoBERTa. The right part is the partially causal masking strategy (white cell: unmasked; grey cell: masked).}
	\label{model_roberta} 
	
\end{figure*}

\paragraph{T5 Version}
The approach based on T5 is described in Section~\ref{4}. We use \texttt{<extra\_id\_0>}, \texttt{<extra\_id\_1>}, \texttt{<extra\_id\_2>}, and \texttt{<extra\_id\_3>} as sentinel tokens \texttt{[X]}, \texttt{[Y]}, \texttt{[Z]}, and \texttt{[W]}.
\paragraph{BART Version}
The model based on BART is basically the same as the T5 version, except that the sentinel tokens in the template are replaced with \texttt{[MASK]} tokens, following the pre-training task format of BART, and the target sequence is composed of entity types and label verbalizations.
\paragraph{RoBERTa Version}
In this section, we detail the implementation based on RoBERTa.
For the RoBERTa version, we concatenate the source and target as inputs together and use partial causal masking to distinguish the encoder-decoder representations,
as shown in Figure~\ref{model_roberta}. Specifically, we utilize one RoBERTa model as both the source and target embeddings by concatenating the source sequence $T(\boldsymbol{x})$ and target $\boldsymbol{y}$ as inputs together. We provide a partial causal attention mask to distinguish the source / target representations. The attention mask – 2-dimensional matrix denotes as $\mathbf{M}$. Specifically, for the source tokens, the mask allows full source self-attention, but mask out all target tokens. For $i =[0, 1, ..., |T(\boldsymbol{x})|-1]$, $|\cdot|$ represents the length of sequence, 

\begin{equation*}
	M_{ij}=
	\begin{cases}
		1, & 0\le j< |T(\boldsymbol{x})|,\\
		0, & otherwise
	\end{cases}
\end{equation*}

To guarantee that the decoder is autoregressive, we enforce the target tokens to only attend to previous tokens and not attend to future tokens. 
For $i =[0, 1, ..., |\boldsymbol{y}|-1] + |T(\boldsymbol{x})|$, 
\begin{equation*}
	M_{ij}=
	\begin{cases}
		1, & 0\le j< |T(\boldsymbol{x})|,\\
		1, & |T(\boldsymbol{x})| \le j \le i, \\
		0, & otherwise
	\end{cases}
\end{equation*}
\begin{table}
	\centering
	\scalebox{0.7}{
		\begin{tabular}{lrrrr}
			\toprule
			Dataset & \#train&\#dev&\#test&\#rel\\
			\midrule
			TACRED&68,124&22,631&15,509&42\\
			TACREV&68,124&22,631&15,509&42\\
			Re-TACRED&58,465&19,584&13,418&40\\
			Wiki80&44,800&5,600&5,600&80\\\bottomrule
		\end{tabular}
	}
	\caption{Statistics of datasets used.}
	\label{data}
\end{table}

\section{Setup and Training Details}\label{C}

The model is implemented with Pytorch\footnote{\url{https://pytorch.org/}} and Huggingface transformer library\footnote{\url{https://github.com/huggingface/transformers}}. T5-large\footnote{\url{https://huggingface.co/t5-large}}, BART-large\footnote{\url{https://huggingface.co/facebook/bart-large}}, and RoBERTa-large\footnote{\url{https://huggingface.co/roberta-large}} are adopted as the backbone to validate our method, which have 770 million, 406 million, and 355 million parameters, respectively.  We utilize 1 NVIDIA Tesla V100 GPU with 32 GB memory to run all experiments. For our implementation based on BART, the training times of TACRED under $K=16$ and fully supervised settings are $0.36$ hours and $10.1$ hours, respectively, and testing time is $0.54$ hours. The maximum generation length $L$ depends on the maximum length of label verbalizations. The length of continuous tokens in the template $T(\cdot)$ is set to $n\times3$, where $n_0=n_1-n_0=n_2-n_1=n$. $n$ is $3$ in our implementation, and detailed discussion is in Section~\ref{5.4}. We use a batch size of $4$ for T5 and $16$ for BART, which are chosen for practical consideration to fit into GPU memory. During training, our model is optimized with AdamW \cite{DBLP:conf/iclr/LoshchilovH19}. The hyper-parameters are manually adjusted based on the performance on the development set. 

\subsection{Low-Resource Setting}
For low-resource setting, only a small amount of labeled data is available, i.e. $K$ labeled instances per relation for training and validation sets. $K$ is chosen from $8$, $16$, and $32$. The model is finally evaluated on the whole test set. To validate the performance under the low-resource setting, we report the average performance across five different randomly sampled training and development data using a fixed set of seeds $S_{seed}=\{13, 21, 42, 87, 100\}$.

The hyper-parameters we chose are shown as follows:
\begin{itemize}
	\item maximum length of the input sequence: $512$
	\item learning rate of optimization for pre-trained language model parameters: $3e-5$
	\item learning rate of optimization for the learnable continuous token in the template: $1e-5$
	\item epochs: $10$ for TACRED, TACREV, Re-TACRED,  and $20$ for Wiki80
\end{itemize}

\subsection{Fully Supervised Setting}
Under this setting, whole training, development, and test data are all available. The hyper-parameters we chose are shown as follows:
\begin{itemize}
	\item maximum length of the input sequence: $512$
	\item learning rate of optimization for pre-trained language model parameters: $3e-5$
	\item learning rate of optimization for the learnable continuous token in the template: $1e-5$
	\item epochs: $5$ for TACRED, TACREV, Re-TACRED,  and $10$ for Wiki80
\end{itemize}
\section{Metrics and Validation Performance}
We use the micro $F_1$ score\footnote{\url{https://scikit-learn.org/stable/modules/generated/sklearn.metrics.f1_score.html}} for test metrics to evaluate models, which is calculated globally by counting the total true positives, false negatives and false positives. The performance of our method on the validation set is shown in Table~\ref{validation}. Note that the validation data under low-resource setting is also obtained by randomly sampling $K$ instances. We report the average micro $F_1$ and standard deviation over 5 different splits.
\begin{table}[h]
	\centering
	\scalebox{0.7}
	{
		\begin{tabular}{lcccc}  
			\toprule
			Dataset &$K$=8&$K$=16&$K$=32&full\\\midrule
			TACRED&80.1 \small{$(\pm2.0)$}&82.7 \small{$(\pm1.9)$}&84.3 \small{$(\pm0.8)$}&74.3\\
			TACREV& 82.4 \small{$(\pm1.1)$} & 85.6 \small{$(\pm1.0)$} & 88.9 \small{$(\pm0.8)$} &82.1\\
			Re-TACRED& 85.6 \small{$(\pm1.8)$} &89.4 \small{$(\pm1.2)$}&91.0 \small{$(\pm0.4)$}&90.8\\
			Wiki80& 82.1 \small{$(\pm0.8)$} & 87.1 \small{$(\pm0.6)$} & 89.1 \small{$(\pm0.2)$} &93.5\\
			\bottomrule
		\end{tabular}
	}
	\caption{
		Micro $F_1$ of GenPT-RoBERTa on the validation set.}
	\label{validation}
\end{table}
\begin{figure*}[t]
	\centering
	\includegraphics[width=\textwidth]{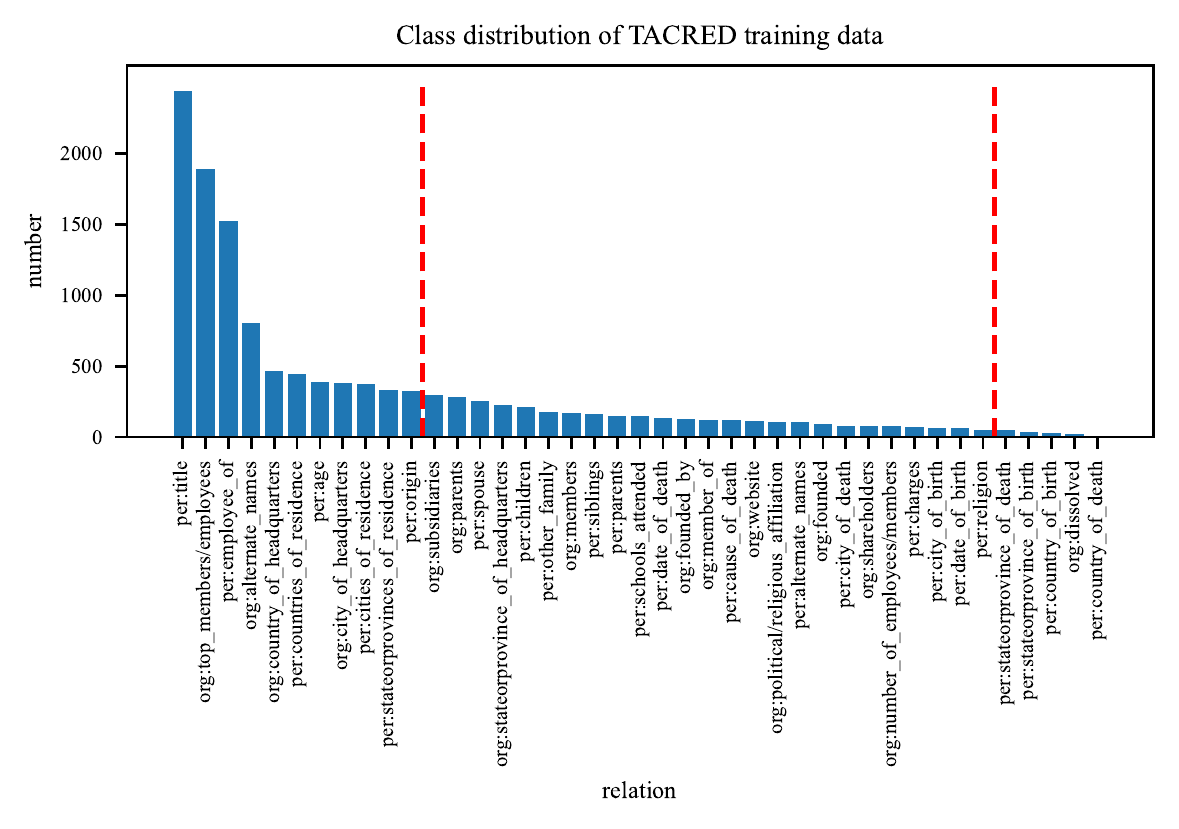}
	\caption{Class distribution of TACRED training data.}
	\label{class_distribution} 
\end{figure*}

\section{Experimental Details}\label{D}
\subsection{Implementation Details of Baselines}
The rest of baseline systems were trained using the hyperparameters reported by the original papers (SpanBERT\footnote{\url{https://github.com/facebookresearch/SpanBERT}}, LUKE\footnote{\url{https://github.com/studio-ousia/luke}}, TANL\footnote{\url{https://github.com/amazon-research/tanl}}, NLI\footnote{\url{https://github.com/osainz59/Ask2Transformers}}, and TYP Marker\footnote{\url{https://github.com/wzhouad/RE_improved_baseline}}). For NLI, we rerun the original code based on RoBERTa-large-mnli\footnote{\url{https://huggingface.co/roberta-large-mnli}}. 
Since they need manual designed template for each relation, we only rerun on the TACRED dataset they use.
We notice that although NLI achieves high average performance, the standard deviation on 5 different training and dev sets is also quite high, as shown in Table ~\ref{nli}. The reason is that they use a threshold-based detection method to detect \textit{no\_relation} class. Performance is highly correlated with the choice of threshold. In addition, NLI takes 0.81 hours during inference, which makes it less efficient than other discriminative models.
\begin{table}[h]
	\centering
	\scalebox{0.7}
	{
		\begin{tabular}{lccc}  
			\toprule
			\multirow{2}*{Model} & 	\multicolumn{3}{c}{TACRED} \\ 
			&$K$=8&$K$=16&$K$=32\\\midrule
			\multirow{2}*{NLI} &31.8&35.6&36.7\\
			&\small{$(\pm2.0)$}&\small{$(\pm5.2)$}&\small{$(\pm3.8)$}\\
			\bottomrule
		\end{tabular}
	}
	\caption{Experimental results of NLI under low-resource setting.}
	\label{nli}
\end{table}

One thing we need to notice is we adopt different few-shot settings compared to the original settings in NLI. Concretely, we follow the work of \citet{gao-etal-2021-making, DBLP:journals/corr/abs-2105-11259, DBLP:journals/corr/abs-2104-07650}, which randomly samples a small set of training and development data, that is, $K$ for each class including \textit{no\_relation}. However, NLI collects data from positive and negative in equal ratio, e.g., under its 5\% split setting, there are around 16 examples per positive relation and 2,756 examples for negative \textit{no\_relation}. The main difference of settings between NLI and us is that they adopt more negative examples during training.
\subsection{Training Relation Frequency}\label{E.2}

As shown in Figure~\ref{class_distribution}, we count the number of training data for each relaiton in TACRED, which follows a long-tailed distribution. To figure out the performance on classes of different frequencies, we split the test data into three subsets according to the class frequency in training. The relations with more than 300 training instances forms a high frequency subset (except for ``\textit{no\_relation}'') 
, those with 50-300 training instances form a middle frequency subset
, and the rest form a low frequency subset. 
The high, middle, and low frequency subsets consist of 11, 25, and 5 relations, with each containing 2,263, 1,024, and 38 instances on TACRED and 2,180, 912, and 31 on TACREV. The specific scores of Figure~\ref{class_freq} are shown in Table~\ref{freq_data}.

\begin{table}[h]
	\centering
	\scalebox{0.7}
	{
		\begin{tabular}{lcccccc}  
			\toprule
			\multirow{2}*{Model} & 	\multicolumn{3}{c}{TACRED}&\multicolumn{3}{c}{TACREV} \\ 
			&high& mid & low&high& mid & low\\\midrule
			TYP Marker &87.0&77.1&40.7&92.4&88.1&53.1\\
			GenPT&87.9&79.8&46.2&93.1&90.9 &61.5\\
			\bottomrule
		\end{tabular}
	}
	\caption{Detailed $F_1$ (\%) scores of different frequency relations on TACRED and TACREV.}
	\label{freq_data}
\end{table}

\begin{table}[t]
	\centering
	\scalebox{0.7}
	{
		\begin{tabular}{lccc}  
			\toprule
			GenPT & TACRED & TACREV & Re-TACRED \\\midrule
			manual& 13.1 &10.3&19.2 \\
            continuous& 15.7 & 12.7 &18.6 \\
			\bottomrule
		\end{tabular}
	}
	\caption{
		Micro $F_1$ (\%) of GenPT (BART) under zero-shot setting.}
	\label{zero-shot}
\end{table}

\subsection{Details of Hand-Crafted Label Words}\label{D.3}
As shown in Table~\ref{handmade-labels}, we list all hand-crafted label words with fixed length (5) for relations in TACRED, which is from PTR\footnote{\url{https://github.com/thunlp/PTR}}. In our experiments, we use non-semantic numeric relation id, partial-semantic handmade verbalization, and full-semantic complete label words to prove the significance of label semantic information for the relation classification task.
\subsection{Details of Likelihood-based Prediction}\label{D.4}
We compare our proposed decoding relation scoring strategy with likelihood-based prediction, which performs prediction using sequence likelihoods as class scores. Specifically, given a pre-defined set of relations $\mathcal{R}$, likelihood-based prediction calculates score for each relation $r$ by feeding the corresponding target sequence $\boldsymbol{y} =\texttt{[X]}\; \boldsymbol{t}_h \; \texttt{[Y]}\; \boldsymbol{t}_t \;\texttt{[Z]}\;  \boldsymbol{r}\; \texttt{[W]} $ to the decoder and taking target sequence likelihoods as class score $s_r$, 
\begin{equation*}
	s_r=\sum_{j=1}^{|\boldsymbol{y}|}\textrm{log}P(y_j|y_{<j}, T(\boldsymbol{x}))
\end{equation*}
where $\boldsymbol{x}$ is the instance to be predicted, $\boldsymbol{t}_h$ and $\boldsymbol{t}_t$ denote the subject entity type and object entity type sequences of $\boldsymbol{x}$, and $\boldsymbol{r}$ represents the token verbalizations of relation $r$. The relation with the highest score is regarded as the predicted class.
\section{Results in Zero-Shot Setting}\label{F}
We conduct experiments with manual template,
\begin{equation*}
\begin{aligned}
	T(\boldsymbol{x})=	\boldsymbol{x}.\, \,  &The\,\,  relation\,\,  between\,\, \texttt{[X]} \,\, \boldsymbol{e}_h\,\,  and\,\,  \\
	&\texttt{[Y]}\,\,  \boldsymbol{e}_t \,\, is \,\, \texttt{[Z]}.
\end{aligned}
\end{equation*}
and continuous prompt (the continuous token length is $0$),
\begin{equation*}
\begin{aligned}
	T(\boldsymbol{x})=\boldsymbol{x}.\, \,  \texttt{[X]} \,\, \boldsymbol{e}_h\,\, \texttt{[Y]}\,\,  \boldsymbol{e}_t \,\, \texttt{[Z]}.
\end{aligned}
\end{equation*}
The experiment results in zero-shot setting are shown in Table~\ref{zero-shot}.

\begin{table*}[t]
	\centering
	\scalebox{0.9}
	{
		\begin{tabular}{lll}  
			\toprule
			Class Label & Hand-made Verbalization&	Relation id  \\\midrule
			per:charges	& person, was, charged, with, event &\texttt{<Rel\_0>} \\
			per:date\_of\_death	&person, was, died, on, date&\texttt{<Rel\_1>} \\
			per:country\_of\_death	&person, was, died, in, country&\texttt{<Rel\_2>} \\
			per:cause\_of\_death		&person,	was, died,	of,	event&\texttt{<Rel\_3>} \\
			org:founded\_by		&organization,	was,	founded,	by,	person&\texttt{<Rel\_4>} \\
			org:founded	&	organization,	was,	founded,	in,	date&\texttt{<Rel\_5>} \\
			per:city\_of\_death&		person,	was, died,	in,	city&\texttt{<Rel\_6>} \\
			per:stateorprovince\_of\_death&		person,	was,	died,	in,	state&\texttt{<Rel\_7>} \\
			per:date\_of\_birth&		person,	was,	born,	in,	date&\texttt{<Rel\_8>} \\
			per:stateorprovince\_of\_birth&		person,	was,	born,	in,	state&\texttt{<Rel\_9>} \\
			per:country\_of\_birth&		person,	was,	born,	in,	country&\texttt{<Rel\_10>} \\
			per:city\_of\_birth	&	person,	was,	born,	in,	city&\texttt{<Rel\_11>} \\
			org:shareholders&		organization,	was,	invested,	by,	person&\texttt{<Rel\_12>} \\
			per:other\_family&		person,	's,	relative,	is,	person&\texttt{<Rel\_13>} \\
			per:title&		person,	's,	title,	is,	title&\texttt{<Rel\_14>} \\
			org:dissolved&		organization,	was,	dissolved,	in,	date&\texttt{<Rel\_15>} \\
			org:stateorprovince\_of\_headquarters&		organization,	was,	located,	in,	state&\texttt{<Rel\_16>} \\
			org:country\_of\_headquarters	&	organization,	was,	located,	in,	country&\texttt{<Rel\_17>} \\
			org:city\_of\_headquarters&		organization,	was,	located,	in,	city&\texttt{<Rel\_18>} \\
			per:countries\_of\_residence&		person,	was,	living,	in,	country&\texttt{<Rel\_19>} \\
			per:stateorprovinces\_of\_residence&		person,	was,	living,	in,	state&\texttt{<Rel\_20>} \\
			per:cities\_of\_residence	&	person,	was,	living,	in,	city&\texttt{<Rel\_21>} \\
			org:member\_of	&	organization,	was,	member,	of,	organization&\texttt{<Rel\_22>} \\
			per:religion&		person,	was,	member,	of,	religion&\texttt{<Rel\_23>} \\
			org:political/religious\_affiliation	&	organization,	was,	member,	of,	religion&\texttt{<Rel\_24>} \\
			org:top\_members/employees&		organization,	's,	employer,	was,	person&\texttt{<Rel\_25>} \\
			org:number\_of\_employees/members	&	organization,	's,	employer,	has,	number&\texttt{<Rel\_26>} \\	
			per:schools\_attended &person,	's,	school,	was, organization&\texttt{<Rel\_27>} \\
			per:employee\_of	&person,	's,	employee,	was,	organization&\texttt{<Rel\_28>} \\
			per:siblings&	person,	's,	sibling,	was	,person&\texttt{<Rel\_29>} \\
			per:spouse	&person,	's,	spouse,	was,	person&\texttt{<Rel\_30>} \\
			per:parents	&person,	's,	parent,	was,	person&\texttt{<Rel\_31>} \\
			per:children&	person,	's,	child,	was,	person&\texttt{<Rel\_32>} \\
			per:alternate\_names	&person,	's,	alias,	was,	person&\texttt{<Rel\_33>} \\
			org:alternate\_names	&organization,	's,	alias,	was,	organization&\texttt{<Rel\_34>} \\
			org:members	&organization,	's,	member,	was,	organization&\texttt{<Rel\_35>} \\
			org:parents	&organization,	's,	parent,	was,	organization&\texttt{<Rel\_36>} \\
			org:subsidiaries&	organization,	's,	subsidiary,	was, organization&\texttt{<Rel\_37>} \\
			per:origin	&person,	's,	nationality,	was, country&\texttt{<Rel\_38>} \\
			org:website	&organization,	's,	website,  was,	url&\texttt{<Rel\_39>} \\
			per:age	&person,	's,	age,	was,	number&\texttt{<Rel\_40>} \\
			no\_relation&	entity,	is,	irrelevant,	to,	entity&\texttt{<Rel\_41>} \\
			\bottomrule
		\end{tabular}
	}
	\caption{Hand-crafted label words with fixed length (5) for relations in TACRED.}
	\label{handmade-labels}
\end{table*}

\end{document}